\documentclass{IEEEtran4PSCC}

\ifCLASSINFOpdf
   \usepackage[pdftex]{graphicx}

\else

\usepackage[dvips]{graphicx}

\fi
 \usepackage{soul}
\usepackage{xcolor}
\usepackage[cmex10]{amsmath}
\usepackage{colortbl}

\usepackage{todonotes}
\usepackage[super]{nth}
\usepackage{hyperref}
 \usepackage{algorithm} 
\usepackage{algpseudocode} 

\makeatletter
\let\old@ps@headings\ps@headings
\let\old@ps@IEEEtitlepagestyle\ps@IEEEtitlepagestyle
\def\psccfooter#1{%
    \def\ps@headings{%
        \old@ps@headings%
        \def\@oddfoot{\strut\hfill#1\hfill\strut}%
        \def\@evenfoot{\strut\hfill#1\hfill\strut}%
    }%
    \def\ps@IEEEtitlepagestyle{%
        \old@ps@IEEEtitlepagestyle%
        \def\@oddfoot{\strut\hfill#1\hfill\strut}%
        \def\@evenfoot{\strut\hfill#1\hfill\strut}%
    }%
    \ps@headings%
}
\makeatother

\psccfooter{%
        \parbox{\textwidth}{\hrulefill \\ \small{22nd Power Systems Computation Conference} \hfill \begin{minipage}{0.2\textwidth}\centering \vspace*{4pt} \includegraphics[scale=0.06]{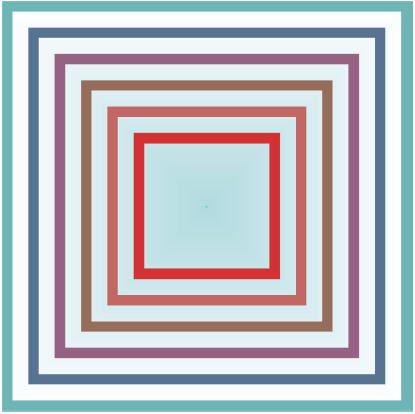}\\\small{PSCC 2022} \end{minipage} \hfill \small{Porto, Portugal --- June 27 -- July 1, 2022}}%
}
\usepackage{enumitem}
\usepackage{caption}
\usepackage{subcaption}
\begin{document}

\title{\vspace{-0.1em} Learning to run a power network with trust \vspace{-0.2em}}

%% To specify the authors when (number of affiliations <= 2)
\author{
\IEEEauthorblockN{Antoine Marot\\ Benjamin Donnot\\Karim Chaouache}
\IEEEauthorblockA{RTE AI Lab,
Paris, France}
\and
\IEEEauthorblockN{Adrian Kelly}
\IEEEauthorblockA{EPRI, Ireland}
\and
\IEEEauthorblockN{Qiuhua Huang}
\IEEEauthorblockA{PNNL, USA}
\and
\IEEEauthorblockN{Ramij-Raja Hossain}
\IEEEauthorblockA{Iowa State University, USA}
\and
\IEEEauthorblockN{Jochen L. Cremer}
\IEEEauthorblockA{TU Delft, Netherlands}}
%% To specify the authors when (number of affiliations > 2)
% \author{\IEEEauthorblockN{Author n.1\IEEEauthorrefmark{1},
% Author n.2\IEEEauthorrefmark{2},
% Author n.3\IEEEauthorrefmark{3}, 
% Author n.4\IEEEauthorrefmark{3} and
% Author n.5\IEEEauthorrefmark{4}}
% \IEEEauthorblockA{\IEEEauthorrefmark{1} Department Name of Organization A\\
% Name of the organization A,
% Address A\\ Emails if wanted}
% \IEEEauthorblockA{\IEEEauthorrefmark{2} Department Name of Organization B\\
% Name of the organization B,
% Address B\\ Emails if wanted}
% \IEEEauthorblockA{\IEEEauthorrefmark{3} Department Name of Organization C\\
% Name of the organization C,
% Address C\\ Emails if wanted}
% \IEEEauthorblockA{\IEEEauthorrefmark{4}Department Name of Organization D\\
% Name of the organization D,
% Address D\\ Emails if wanted}
% }
\maketitle

\begin{abstract}

Artificial agents are promising for real-time power network operations, particularly, to compute remedial actions for
congestion management. However, due to high reliability requirements, purely autonomous agents will not be deployed any time soon and operators will be in charge of taking action for the foreseeable  future. Aiming at designing assistant for operators, we instead consider humans in the loop and propose an original formulation. We first
advance an agent with the ability to send to the operator alarms
ahead of time when the proposed actions are of low confidence.
We further model the operator’s available attention as a budget
that decreases when alarms are sent. We present the design and
results of our competition ”Learning to run a power network with
trust” in which we evaluate our formulation and benchmark the ability of submitted agents to send relevant alarms while operating the network to their best.
\end{abstract}


\section{Introduction}
Power network operators are in charge of maintaining a reliable, secure supply of electricity at all times. A vast majority of the real-time operation and control decisions are made by human operators based on their experiences, and predefined operation rules and manuals. However, real-time decision-making is getting more challenging as the human operator has to deal with more information, more uncertainty, more applications and more coordination \cite{marot2022perspectives}. Recent power outages such as the Texas power outages in early 2021 clearly showed that human operators faced daunting challenges in dealing with rare events and they desperately need intelligent decision-support tools to help make fast and robust decisions to safeguard the network. The ability to foresee events ahead of time is also vital to the future operation of the power system, given inherent future variability. 

%Their task is becoming increasingly difficult as networks are becoming more complex. More renewable energy sources, storage and flexible demand are being integrated into the network. Renewable generation and variable demand add uncertainty ahead of time and flexible storage adds dependencies from one time period to another. The devices on the network are also becoming digitised, offering more control capabilities, bringing more data to analyse, and more actions to choose from. Operators now have massive amounts of data but are required to make complex decisions and coordinated actions, very close to real-time. Additionally, the task of finding coordinated, optimal actions involves novel constraints in time. 

%Network operators must coordinate with neighbouring network operators, distribution system operators and market participants to maintain reliability and security at all times (also known as reliability coordination). 
%The vision of this research is that the human operator will remain in charge of the system. Reliability management is a critical task and the responsibility currently lies with humans in the control centre. In the future, the human may supervise automation, with artificial agents as assistants, monitoring the current system and projecting the forecasted system via simulation proposing actions to the operator when issues are identified \cite{marot2020towards}.

Human operators and AI can be seen as complementary heterogeneous intelligence that could achieve a superior outcome when combined\cite{dellermann2019hybrid}. In the future, the human may supervise automation, with artificial agents as so-called assistants, monitoring the current system and projecting the forecasted system via simulation. The assistants may propose actions to the operator when issues are identified \cite{marot2020towards}, ultimately having good foresight to securely operate the system.

Machine Learning (ML) and Reinforcement Learning (RL) models are showing promise for managing operational reliability \cite{Weh20, Gla17,Yan21}. ML and RL can propose operating control decisions very quickly, making it suitable for emergency control purposes \cite{Hua19}. %Tailored, automated (expert) systems can discover and propose efficient control actions beyond what has been observed in the pas \cite{marot2018expert}. Beyond tailored solutions, 
Autonomous agents trained with RL are particularly promising as they can reinforce its leanings, even on very complex tasks. Hence, an agent can autonomously improve itself with training simulations, just as human operators adapted their heuristics for their network with experience over years and training on simulated scenarios. It has previously been shown that RL based agents can autonomously improve its own model quality %, for instance, for optimal operation and maintenance \cite{Roc19}
for real-time power network operation management, through the "Learning to run a power network" (L2RPN) competition series.
%and operational congestion management
\cite{marot2021learning}.
\begin{figure}
\centering
\includegraphics[width=\linewidth]{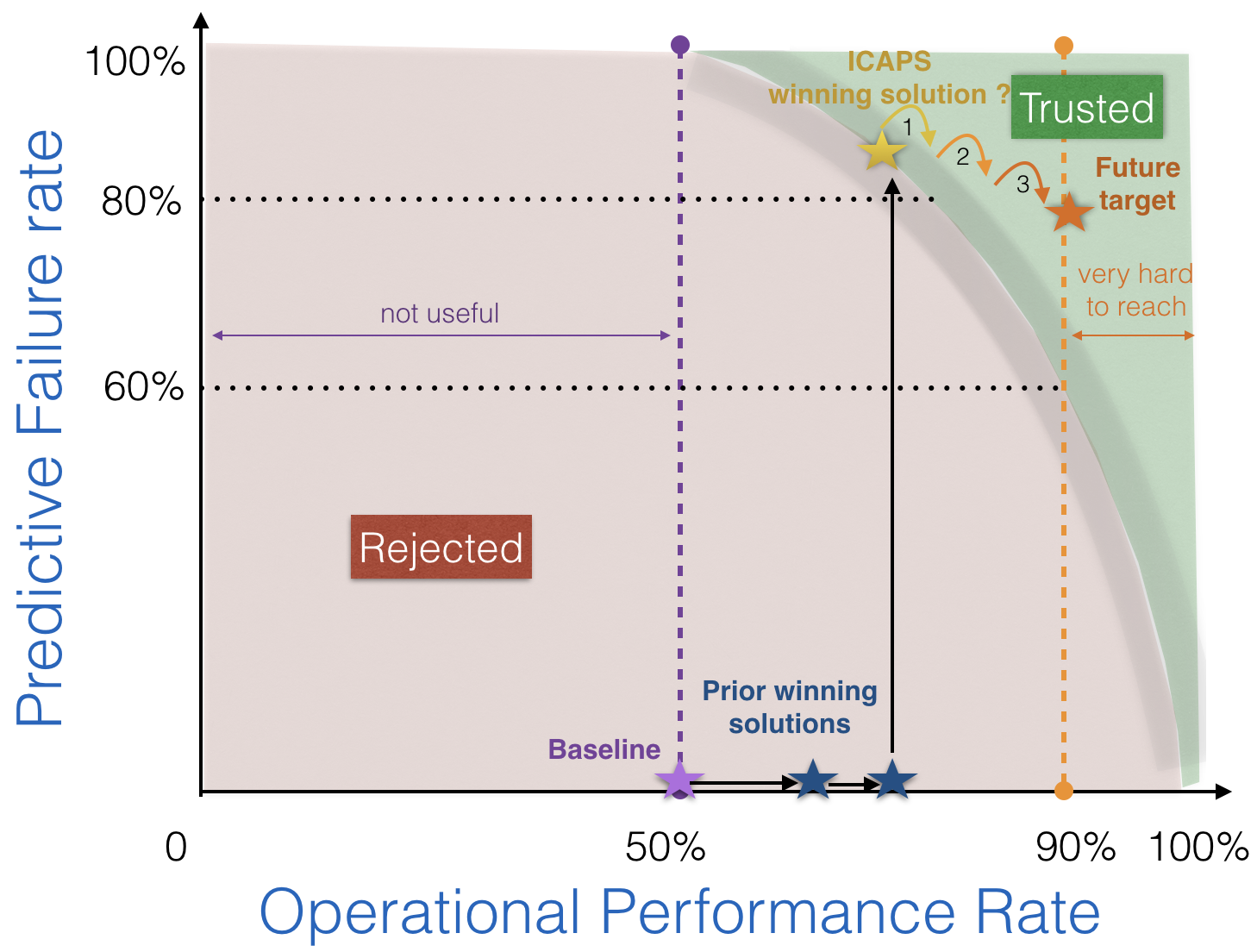}
%\caption{To build trust, communication of anticipated risks to the operator is needed in addition to overall operational agent performance}
\caption{
As a first approximation, trust in an agent can emerge under appropriate levels of operational performance and predict consequence rates. Here we show the expected path of successive L2RPN winning solutions to realistically develop a trusted agent when adding a predict failure feature.}
\label{fig:fig_trust_performance}
\end{figure}
Starting from our initial baseline \cite{marot2018expert}, winning solutions of these successive L2RPN competitions have progressively improved the operational performance of artificial agents to robustly operate (even under N-1 line disconnections) the network \cite{lan2020ai},\cite{yoon2020winning},\cite{zhou2021action} as illustrated on Figure \ref{fig:fig_trust_performance}  along the x-axis.

%For such complex tasks, agents already outperform in some cases humans in the abilities to digest large amounts of data and to select actions for large, complex tasks. This autonomous adaptability seems an important feature of managing a constantly changing network. Yet, even the  best  agents  still  failed  over  30\% of  the  L2RPN test  scenarios, which we can deem as catastrophic failures since no prior warning is sent. Hence the operator would quickly lose confidence and trust in such agent that he would have to supervise continuously: he will reject such assistant. While a minimum operational performance is necessary for an assistant to be useful, an almost perfect performance level is probably out of reach. Therefore, another dimension needs to be considered to build trustworthy agent.

However, existing AI technology lacks the robustness and trustworthiness that are required for high-consequence, high impact, decision-making in real-time network operation. One key issue is insufficient consideration of leveraging the experience of and working with human teammates (or operators) in existing AI models and applications. Experienced operators in power network deploy extensive domain or expert knowledge that cannot adequately represented mathematically or easily captured by existing machine learning models\cite{wilson2015evaluation}. A recent study by MIT researchers showed that state-of-the-art AI agents can become frustrating teammates, and highlighted the need of incorporating subjective metrics such as trust and teamwork into the development of assistants\cite{siu2021evaluation} . 

%link between ML, vision and trust
%The vision of this research %is that the human operator %remains in charge of the %system. Reliability %management is a critical task %and the responsibility %currently lies with humans in %the control centre. In the %future, the human may %supervise automation, with RL %agents as assistants, %monitoring the current system %and projecting the forcasted %system via simulation %proposing actions to the %operator when issues are %identified %\cite{marot2020towards}. However, if ML/RL-based and methods are deployed in the control centre, humans will require a high level of trust in these agents to be effective in real-time. \cite{cho2015survey}. 

%\adrian{We should move the below paragraph to the section above describing trust, its out of place here. If not remove altogether}
Research suggests that an AI agent can increase its trustworthiness by reducing conflicting evidence and by increasing the amount of evidence it has gathered\cite{wang2010evidence, Brun20}.Therefore, based on an imperfect and reinforced model, the assistant proposes actions with varying confidence to reduce conflicting evidence. This is represented through the agent Predictive Failure rate dimension in Figure \ref{fig:fig_trust_performance}. This representation makes the low confidence of agents explicit. Working along that direction could eventually make an imperfect agent trustworthy, as the operator will know when to take over. It will also relieve the operator from constant supervision, (hyper-vigilance) which might by physiologically impossible. This also relates to 2-D Hybrid Intelligence diagram target\cite{shneiderman2020human} which represents simultaneously high levels of automation and yet high level of human control. %How this formulation can be achieved in the power system operation context is not yet clear. 

The objective of this paper is to develop an original formulation that incorporates trust-building mechanisms into the process of intelligent assistant learning to securely run the network against various network overloading and physical violation conditions. The framework was tested and demonstrated in the \textit{L2RPN with trust} competition which was organised over the summer 2021. It should be noted that developing an efficient and effective sequential decision making (SDM) formulation is not trivial, but rather critical for obtaining a competent and trustworthy AI assistant for network operators. This is analogous to a novel, complex formulation of the well-known (optimal power flow) OPF problem in power systems\cite{low2014Convex}, close to SCOPF (security constrained OPF) \cite{bhaskar2011security} or Multi-period AC SCOPF \cite{alizadeh2021toward} in particular, but with a human consideration, built in. 

The specific contributions of this paper are:
\begin{enumerate}[label=(\roman*)]
\item proposing a novel SDM formulation in Section. \ref{sec:trust} that incorporates human-AI trust-building mechanisms in the design of intelligent assistants. Agents are now given the ability to send interpretable warning to the human - modelled using novel "attention budget" constraints. 
\item instantiating this concept through the \textit{L2RPN with trust} framework environment in Section. \ref{sec:design}
\item analysing an open competition results to evaluate this concept design in Section. \ref{sec:result} and promising directions.
\end{enumerate}  

\begin{figure*}
\centering
\begin{subfigure}{.4\textwidth}
  \centering
  \includegraphics[width=.8\linewidth]{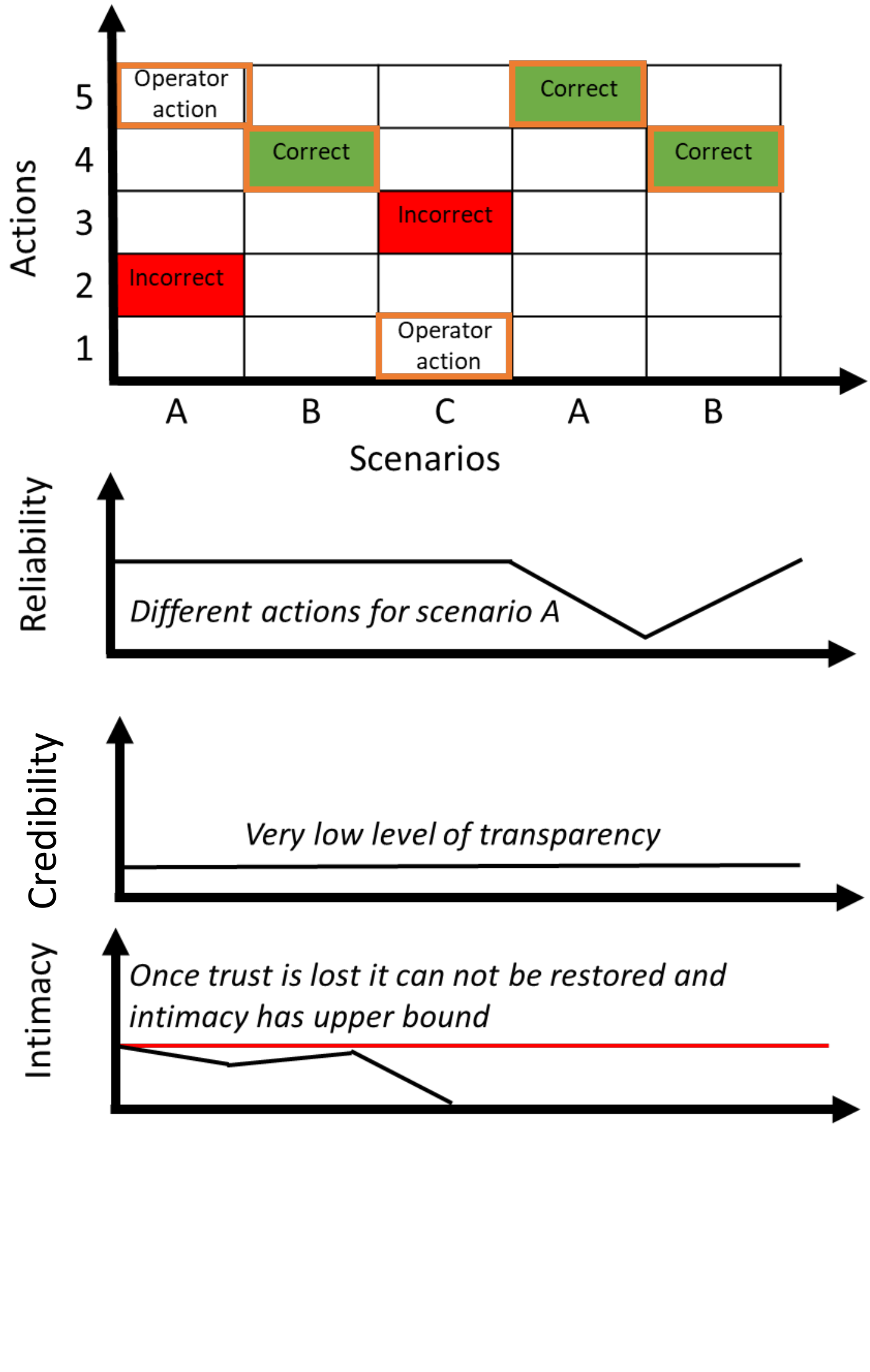}
  \caption{Standard agent}
  \label{fig:withoutrust}
\end{subfigure}%
\begin{subfigure}{.4\textwidth}
  \centering
  \includegraphics[width=.8\linewidth]{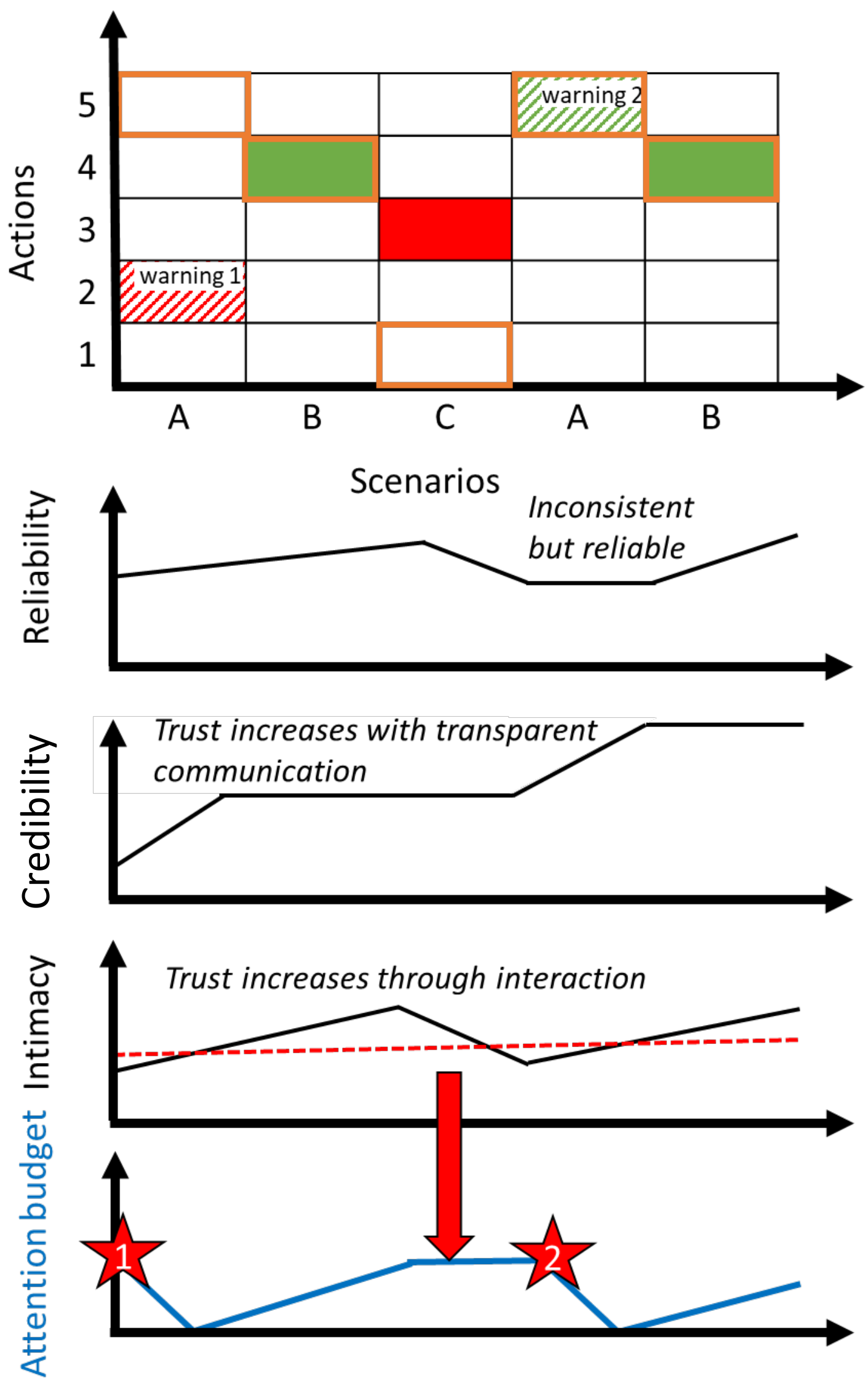}
  \caption{Proposed approach}
  \label{fig:withtrust}
\end{subfigure}%
\caption{Trust concept and the proposed model using attention budget and warnings for actions. The proposed approach considers attention budget of the human, a warning function, and to explain about the region increasing credibility. }
\label{fig:trustconcept}
\end{figure*}

  %most of the time making a complex problem with heuristics as simple as possible (but inefficient). 

\section{Trust in Artificial Intelligence (AI)}\label{sec:trust}
There are inherent issues with automation of tasks more generally \cite{BainbridgeAutomation}, when agents are deployed as assistants to achieve higher efficiencies in managing complex systems. Trust between the human and agent will be difficult to achieve at first as it can not be varied and be completely lost \cite{Jac21}. Therefore, it seems promising to investigate the very fundamental concept of trust within humans (in this case, operators).  This paper investigates whether human operators can develop trust in RL agents to address the issue of missing trust and rigorousness, which currently represents a barrier to their deployment. This idea connects to a broader topic as trustworthiness of AI which is generally believed to be a must-have property for mission-critical applications such as reliability management, in particular context such as vehicle driving or network operations.

The assumption for the proposed trust concept is that humans will trust an agent if they believe that the agent will act in the human's best interest, and accepts vulnerability to the agent's actions (which is adapted from the basic definition of trust \cite{May95}). Before a human can trust (an agent), high levels of (i) credibility, (ii) reliability and (iii) intimacy are required according to the Trust Equation (by Charles Green):

\begin{enumerate}[label=\roman*)]
\item the credibility of an agent can increase when the agent is transparent and explains the proposed actions \cite{Bar20}. %In the context of this work, 
Credibility is an output of increasing transparency, however, transparency is not always necessary for credibility in the extreme, idealistic case of a perfect agent - for instance. Although trustworthiness should be a property of any explainable model, not every trustworthy model is explainable on its own. As an example, for emergency network control, \cite{Zhang2020_XRL} explains RL actions by providing the human with a series of 
summary plots. %offering explanations for the actions. 
%Credibility has to do with transparency and explainability. Explainable AI \cite{Bar20} and its extensions and applications in RL \cite{Zhang2020_XRL} are under active development recently. It should be noted that not every trustworthy model can be explainable on its own, although trustworthiness should most certainly be a property of any explainable model \cite{Bar20}.
%\adrian{Ive moved Reliability to ii since we list it in the set up as ii}
\item trusting an agent requires reliability of the actions. 
%The reliability is another important aspect for human to trust the RL agent. 
A reliable AI agent should work consistently for the same or similar scenarios that it 'sees' during training with a strong generalisation capability and 'know' the limit of its capabilities. There are two approaches that can be used to quantify the limits of an agent and algorithm, passive or active. In the passive approach, a level of confidence is quantified for each suggested automated action/prediction \cite{tetreault2007estimating}, and the user can act accordingly. The more active approach is to receive a signal of 'low confidence' to actively warn the user. While the nature of the information is the same, the confidence of a proposed action, may have a different impact on building trust between humans and agents. The active approach is usually utilised in automated driving of cars, where the autonomous agent warns the driver to take over under some perceived emergency conditions \cite{Vis18}.

\item developing intimacy with an agent is needed. Similar to humans, where intimacy grows with the length of a relationship,
%\adrian{not sure that is true in all cases - I would say confidence and reliability grow over time but intimacy wanes over time. Are we defining it right?}
the life-cycle of an RL can be considered as a whole. For instance, trust, when lost, is difficult to restore. \cite{Toreini2020} identified how trust can be enhanced in the various stages of an AI-based system’s life-cycle, specifically in the design, development and deployment stages, and introduced the concept of an AI Chain of Trust to discuss the various stages and their interrelations. 
%and the only comparison could be drawn to the timing For the intimacy aspect of RL agents, it is like human building up a relationship which takes time. 
%Thus, trust can be impacted through the life-cycle of RL agents (RL-based decision system)\cite{Toreini2020}. From this perspective, both credibility and reliability are also important for winning human's long-term trust. 
\end{enumerate}

Trust between humans and agents relates to these three aspects; reliability, credibility, and intimacy. Unfortunately, standard, or sub-optimally designed AI agents result in low levels of trust build-up as illustrated in Fig. \ref{fig:withoutrust}. This illustrative example shows sequential decision making where the agent proposes exactly $1$ out of $5$ different actions in each sequential scenario. The operator considers the proposed action but may decide, in some cases, on a different action based on other tools, or experience. Therefore, sometimes the agent may propose an incorrect action in conflict with the operators' expectation, and in that case, the intimacy may decrease, and, as no explanation for incorrect actions is provided, the credibility stays at low levels. 
Sometimes, however the agent can “surprise” and teach the human through the agent's proposed actions that humans would normally not take. There, the human would approve this new action despite it being beyond the human’s experience and having not trained for it. If successful, this new proposed action can then become a new strategy that humans will take in the future, and it becomes part of the human operational experience. However, as no interaction further considers the operator mental state, the operator can never fully trust the standard agent as the minimum level (red line) of intimacy is never surpassed. The reliability of the agent may improve with the experience of the agent which is to propose consistently the correct action in the correct scenario. 
In this illustration, the agent proposes two different actions in the scenarios, resulting in reliability decreases because of this inconsistency. 

%We believe a more straightforward approach, particularly for human-agent interactions, is to provide 
%warning or signals which are send in advance to human to show when the agent is going to fail. Sending a signal to indicate limitations is similar to the autopilot (auto-driving) system of cars asking the human driver to take over under some perceived emergency conditions \cite{Vis18}. %This paper will focus on this aspect to enhance trustworthiness of RL for human-in-the-loop grid operation. 

The proposed concept for human / agent interaction aims at improving trust, considering all three aspects of the model: credibility, reliability and intimacy. These three aspects are modelled as an attention budget of the human and warning signals from the agent to the human. As illustrated in Fig. \ref{fig:withtrust}, the agent can actively send warning signals to the human when the agent's confidence about its own actions is low. Sending these warning signals improves reliability, as well as credibility when it provides selective enough details. The warning signals can be discrete, continuous information about the confidence or aiming at explaining the warnings (e.g., in this challenge regional signals are supplied to further improve credibility of agents). The attention budget develops over time (similar as intimacy). The attention budget decreases, when the agent warns the human. %(to not give attention to the agent). 
Intimacy can increase if the warning was relevant or else it will decrease. In case of unwarned failure while the operator could have paid attention, intimacy is modelled to decreases substantially. The attention budget is a balance for operators to decide when they can trust (the agent) or %when to trust 
their own experience. A more accurate, and transparent agent will build trust and will result in overall higher available attention and reduced supervision requirements.

\section{A New Competition Design for Human AI-agent Trust-building} \label{sec:design}

Following the trust concept and model described above, we developed a new L2RPN competition in 2021 with the trust-building between human operator and AI-agent as the focus. The competition was organised through the Codalab platform in Summer-Autumn 2021, as part of the ICAPS conference (International Conference on Automated Planning and Scheduling) and attracted 100 participants. An overview of the competition is provided below, followed by more details.

\subsection{Competition Overview}

Besides operational performance, the L2RPN 2021 competition is structured to build trust between humans and agents using the credibility, reliability, intimacy framework. Entrants are encouraged to design their agents to grade how confident it is of achieving a positive outcome (reward) for an action. It should send an alarm (to the operator) when the proposed actions are of low confidence. This is a proxy for identification of upcoming cascade failure and serves to reduce the conflict in evidence for the human operator (\textbf{reliability}), and to alert them to impending system issues. When formulating the problem, the issue of over-alarming was a risk to positive human-agent interaction. Conversely, the human operator supervising automated systems can experience "too much reliability", otherwise known as  "out-of-the-loop" effect if the human is not warned or given enough time to respond. 
This is where operators are cognitively dis-engaged from real time monitoring and control. When forced to intervene, they are not aware of what or where the problem is. Both illustrates the need for the agent to consider the operator's state in its interactions (\textbf{intimacy}): the relationship quality depends on the right level of solicitations. %or a "boy-who-cried-wolf" effect of alarming spuriously or too often. 
We propose to model a budget for the operator's available attention so that the agent can be designed to consider the human in-the-loop, and which incentivizes the agent to choose the best times for interactions under the attention constraints. 
%the operator's available attention was modelled using a budget that decreases when alarms are sent and returns to maximum incrementally over a medium-long time horizon (intimacy). 
%A further proxy for operator trust building was the ability to inhibit the agent below a user-specified trust threshold. So if the agent registered confidence below, say 50\%, it is prevented from acting autonomously and control shifts to the operator, for manual intervention. 
%Human operator supervising automated systems can experience "too much reliability", namely  "out-of-the-loop" effect. %This is where operators are cognitively dis-engaged from real time monitoring and control and when forced to intervene, they are not aware of what or where the problem is. This can add time to problem identification and solution development, especially when time is very limited. On large electricity networks this is a particular challenge, since issues are localised initially, but can cascade. 
Finally, agents are incentivized to selectively explain when and where a problem originated among pre-defined network areas (\textbf{credibility}).
 
%Intimacy is modelled in the environment by having the agent indicate where the issue is originating .
%These innovations will help build trust between the human and agent by 1) Attracting the operator's attention 2) Directing the operator's attention. 
%\antoine{dealing with "operator's attention budget" and know when it can do less supervision}

The participants were eventually evaluated on a score computed over 24 5-minute resolution weekly scenarios. It was composed of the alarm score (detailed after) and the network operation cost score (see \cite{marot2020l2rpn}) with the following weighting:
\begin{equation}
    Score=0.3*{Score}_{Alarm}+0.7*{Score}_{OperationCost}
\end{equation}

\subsection{Power network operation environment}
\begin{figure}[htbp]
\centering
\includegraphics[width=\linewidth]{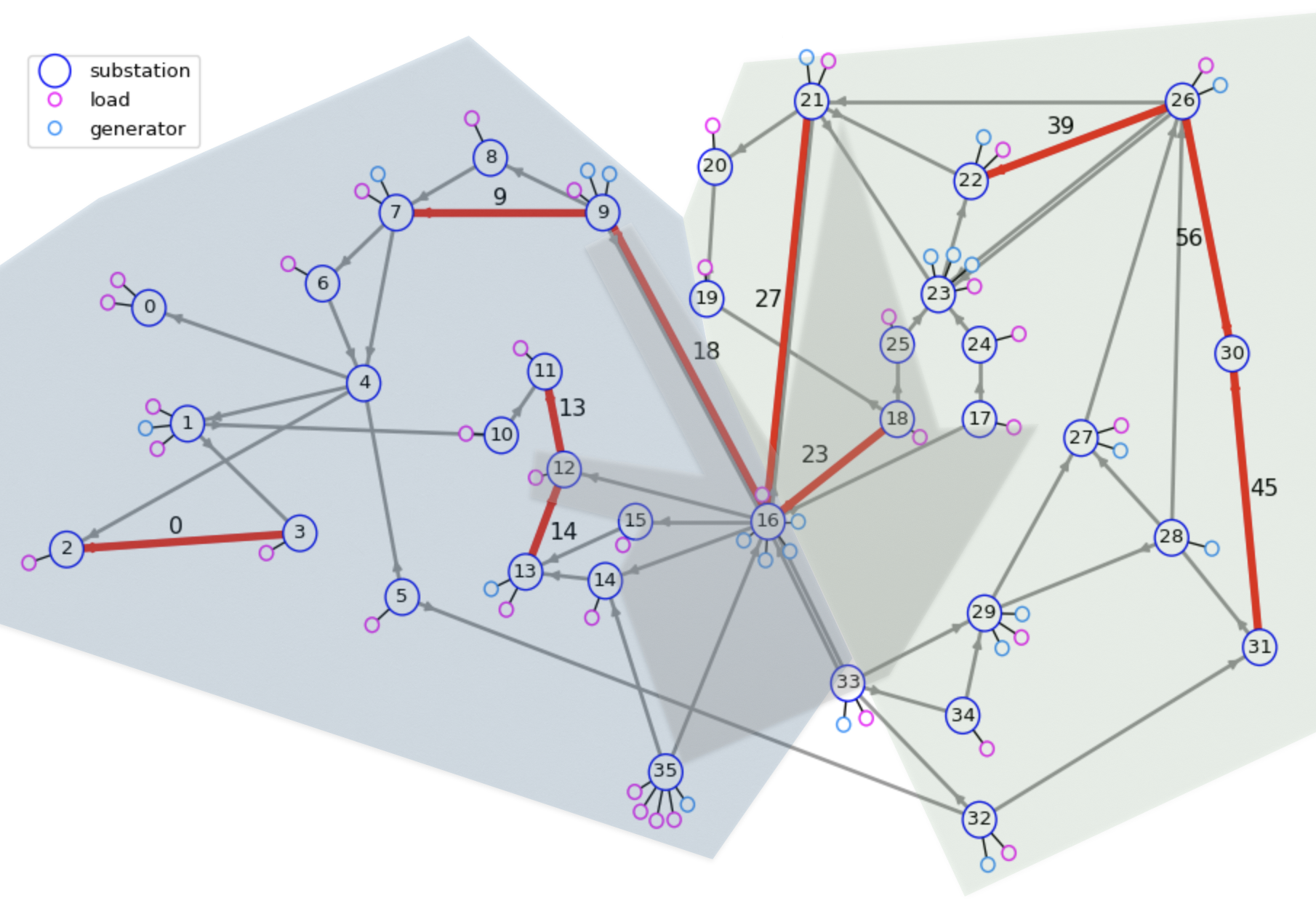}
\caption{Top right IEEE 118 subgrid. Attackable lines by the opponent are red colored. 3 alarm regions are highlighted. }

\label{fig:fig_grid}
\end{figure}

The competition is based on one third subgrid of IEEE 118-bus system as in \cite{marot2021learning} and showed on Figure \ref{fig:fig_grid}.

\begin{figure}[htbp]
\centering
\includegraphics[width=\linewidth]{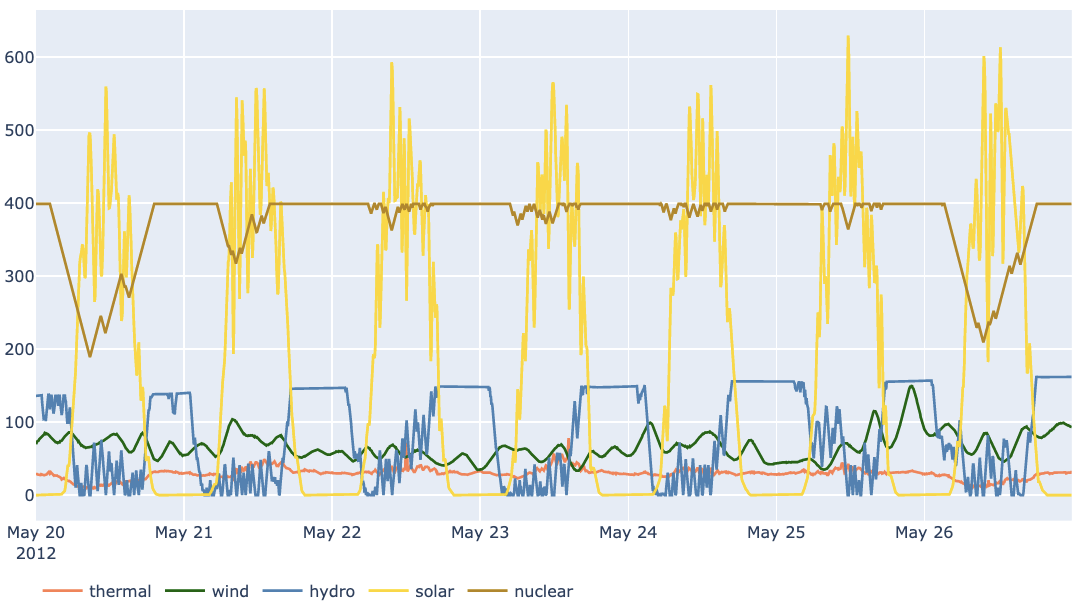}
\caption{Weekly production example per carrier type in May}

\label{fig:fig_timeserie}
\end{figure}

The renewable share makes up to 20\% of the overall energy mix, which is a proxy for high variability in network operation parameters. Monthly Production and Load consumption with a 5-minute resolution time was made available in the training environment and are representative every month of the year (see example in figure \ref{fig:fig_timeserie}). They have been generated through the open-source Chronix2grid package \footnote{see \url{https://github.com/BDonnot/ChroniX2Grid/tree/master/input_data/generation/case118_l2rpn_icaps_2x}}. 

%\antoine{should we give statistics about overload frequency here ? }

%\antoine{should we have this high level MDP description in section 2 rather ? }
The L2RPN Markov Decision Process (MDP) formulation have been previously described here \cite{marot2020learning} and is implemented in Grid2op \cite{grid2op}. The most notable details about the environment, observation and action spaces are described below. 
The agents can take actions considering the following constraints:
    
\begin{itemize}
    %\item	Producers or consumers are connected at various nodes of the graph.  
    \item	Events such as maintenance (deterministic) and line disconnections (stochastic and adversarial) can disconnect power lines for several hours. 
   \item	Power lines have thermal limits. If a power line is overloaded for too long(e.g 15 minutes), it automatically disconnects. This can lead to cascading failures.
    \item An agent must wait few hours before reconnecting an incurred line disconnection (e.g one day).% powerline and must wait specific time period to elapse before doing so. 
    \item Additionally, to avoid expensive network asset degradation or failure, an agent cannot act too frequently on the same asset (e.g not more than once in a 15-minute time period) or perform too many actions at the same time (e.g topological action over 1 substation per step).
 
\end{itemize}
The "Game Over" condition is triggered if total demand is not met any more, that is if some consumption is lost at a substation, possibly because of a cascading failure. 
About the action space, possible actions are:
\begin{itemize}
\item Cheap topology changes (discrete and combinatorial actions) that allow for line disconnection/re-connection and substation nodal re-configurations.
\item Costly production changes (continuous actions) through redispatching or now curtailment. They can be modified within the physical constraints of each plant over time. 

\end{itemize}
The final action space has more than $70,000$ discrete actions and a $20$-dimensional continuous action space. 
    
Considering the observation space, agents can observe complete states of the power network at every step. This includes flows on each power line, electricity consumed and produced at each node, power line disconnection duration etc. 
After verification of the previously described constraints, each action is fed into an underlying power flow simulator to run AC powerflow \cite{lightsim2grid} to calculate the next network state. Agent also have the opportunity to \textbf{simulate} one's action effect on the current state, to validate its action for instance as a human operator would do.
But the future remains unknown: anticipating contingencies is not possible, upcoming productions and consumption are stochastic. 
%\antoine{end of high level MDP description}

The novelty of this competition environment comes from the consideration and interaction of three different kinds of "agents" within the environment: AI based agent, the human operator which needs to focus its attention when it's the most important, and an opponent which emulates contingencies that the network must be robust against. We will hence describe them in the following dedicated subsections.

\subsection{Alarm and operator's attention modelling}
% \antoine{add the equations}
%\antoine{should talk about the calibration and choice of zones}
%\benjamin{TODO: add back the sentence that says that alarm are new type of actions. That has no impact on the environment (the network) itself, but only on the operator}
An agent should be designed to send alarms at a given time while specifying an area among 3 pre-defined ones (as in Figure \ref{fig:fig_grid}). This area demarcation does not have a direct effect on the environment, but will enable desired interactions with an operator, who might modify it, based on the information from the agent. 

%We wanted to model the operator attention in such a way that it was compatible with the Markov Decision Process that models the environment dynamics while being representative of the problem at stake: operator should not be solicited too much nor too often. However, when a problem occur on the network, the operator should have been warned beforehand.

With regard to the operator's attention, we model it as  an "attention budget" $\alpha_t$ at each step $t$, compatible with an MDP formulation. Each time an agent tries to raise an alarm to require the human attention, it has a cost of $\kappa$ (held constant and set to $\kappa = 1$). On the other side, if the agent does not require the operator attention, then the "attention budget" increases by $\mu > 0$ (1.5 per day or per 288 timesteps here). Then, we model the operator attention as:
\begin{enumerate}
    \item $\alpha_{t+1} = \alpha_{t} - \kappa$ if an agent raised an alarm
    \item $\alpha_{t+1} = \alpha_{t} + \mu$ otherwise

\end{enumerate}

Human attention is limited in reality. To make sure that an agent cannot raise alarms too often, the attention budget $\alpha_t$ is capped to a maximum value $A$ (A=3 here). This ensures that the agent cannot raise more than $\frac{A}{\kappa}$
consecutive alarms. Indeed, it can only raise an alarm if the attention budget is above cost $\kappa$. Otherwise it has to wait to recover the necessary budget.

%To ensure that the agent cannot raise too many alarms, we also impose that $\alpha_t >= 0$. This entails that for each step $t$, if $\alpha_t < \kappa$ then the operator will not receive the alarm sent by the agent. This is equivalent to the situation where the agent did not send any alarm. Hence, the agents must carefully chose at which step to send their alarm: for example, if $\alpha_t = 0$ they need to wait $\frac{\mu}{\kappa}$ steps before being able to send another alarm.

In case of failure at timestep $\bar{t}$, an operator should ideally be warned $T_{\text{opt}} = 35$ minutes ahead of time to make a more complex study and take an informed decision. An alarm is considered relevant if sent within $[\bar{t} - (T_{\text{opt}} + T_{\text{width}}),\bar{t} - (T_{\text{opt}} - T_{\text{width}})]$, with $T_{\text{width}} =$ 25 minutes here. An alarm will hence not be considered if raised in the final 10 minutes, before a blackout event, as it is too late for a human operator to perform a study in response to the alarm. An alarm sent greater than one hour is not considered either as this is not selective enough. 35 minutes was chosen as optimal, but may be adjusted for future competitions.

Finally, an alarm score function $\bar{S}$ rewards the agent for sending proper alarms at the right time ahead of failure:
\begin{enumerate}
    \item if no failure occurs,  $\bar{S}$ is given it maximum value, 100 points here, as avoiding failure should always be favored.
    \item if the agent fails the scenario at $\bar{t}$ but raised an relevant alarm at $t_a$ then $\bar{S}=100 \left(1 - \frac{\left|T_{\text{opt}} - \left| \bar{t}-t_a \right| \right| }{ T_{\text{width}}} \right)\times F_{area}$
    \item else if the agent failed to raise an alarm and the system blacks out, it gets a penalty score of -200 points.

\end{enumerate}
$F_{area}$ is a multiplying factor depending on if the alarm spotted the right area of cascade ($F_{area}=$1) or not ($F_{area}=$0.67).
\textbf{NB.} If an agent sends valid alarms at different times $t_a$, the maximum score of each of the valid alarms is taken.

\subsection{Opponent modelling and considerations}

The strategies implemented by the agents in the competition must be robust to unexpected network events, whether natural or intentional. To promote this robustness, we have kept the adversarial approach \cite{omnes2020adversarial} again for the 2021 competition. We have placed in the environment a “special agent” - an “opponent” - whose role is to simulate failures on the network at particular times. the agent must respond to this adversarial attacks on the network.

Three principles are important in the opponent design:

\begin{itemize}
\item Aggressiveness: A too aggressive opponent can bias the competition towards some kind of unrealistic game far from operational concerns. It can also discourage people from participating in the competition.

\item Unpredictability: It is also important for the opponent to be as unpredictable as possible, since we do not want the agents to learn and predict the behaviour of the opponent and adapt specifically to it.

\item Fairness between the participants. The opponent must present the same aggressiveness to all participants.

\end{itemize}

A few improvements have been made for this edition:
\begin{itemize}
\item Attack times: These are random. For more unpredictability, they are drawn according to an exponential distribution (geometric distribution in discrete time) calibrated to have roughly one attack per day on average but not always exactly one per day as before.

\item Durations of the attacks: These are changing following an exponential distribution (they were fixed to four hours in the previous edition) as seen in Figure \ref{fig:fig_behaviour} but with a within a duration constraint of 2 to 8 hours. %They are also based on exponential distributions. 

\item Attacked lines. In order to reflect the idea that the most electrically loaded lines are generally the most prone to failures, we have weighted the probability for a line of being the object of the current attack by the load factor of the line. On average this year, some lines get more attacked than others, but within a maximum 1:4 ratio from the most attacked one to the least attacked one.

%The only thing left to do is to choose the line to attack. The  In order to reflect the idea that the most electrically loaded lines are generally the most prone to failures, we have weighted the probability for a line of being the object of the current attack by the load factor of the line. To prevent some lightly loaded lines from ever being attacked, the way we weight the probabilities ensures a non-zero minimum probability of being attacked for each (attackable) line.

\end{itemize}

%\begin{figure}[htbp]
%\centering
%\includegraphics[width=\linewid%th]{pscc2022_template_LaTeX/pic%tures/attack_durations.png}
%\caption{Attack durations are %random in this new challenge %edition (test sample)}
%
%\label{fig:fig_attack_durations%}
%\end{figure}

%Info : Test sample statistics : 
%158 attacks (in 24 episodes of 7 days)
%| mean(attack duration) = 3 h 52 min %(close to 4 hours)
%| median(attack duration) = 3 h 35 min
%| sd(attack duration) = 1 h 26 min
%| max(attack duration) = 7 h 35 min

%\break
This year again, to avoid having too aggressive attacks, we have kept the principle of one attacked electric line at a time. No multiple attacks. The 10 same attacked lines are shown on Figure \ref{fig:fig_grid}.
It is important to note that for fairness the attack times and durations are the same for everyone in the evaluation scenarios (even if these times and durations are unknown to the participants), but not necessarily attacks on the same lines.

%\begin{figure}[htbp]
%\centering
%\includegraphics[width=\linewid%th]{pscc2022_template_LaTeX/pic%tures/steps_lines_under_attack.%png}
%\caption{Lines are attacked %according to their load factor %(test sample)}
%
%\label{fig:fig_steps_lines_atta%cked}
%\end{figure}

\section{Evaluation of competition design}\label{sec:result}
%\antoine{different kind of possible results we could aim at:3) train an agent with and without the alarm. Does it change it operational behavior ? 5) Use topological oracle to get some feasibility and score bounds ?}

This section evaluates the designed competition with trust concept by analysing the results of the competition and further investigating simple example agents.
%We will now analyse the results of the agents, describing when they performed best and worse. \textcolor{blue}{Through this large and open experiment, we also evaluate the effectiveness of our trust design and calibration}. We will eventually try to find out whether the best operating agents also show to be most aware of their limits, and therefore the most suitable assistant or not. 
%A mix of expert system and learning agents were submitted for this first challenge. Submitted expert system mainly tried to deal with overflows and most of the time finished the scenarios, but hardly optimized the network otherwise. Learning agent had a harder time at first to be robust to overflows but eventually managed them, and in addition they optimized the network capacity continuously. They also showed the use of more complex and coordinated actions at the end that we will describe. 

% Use this to place sponsorships

\subsection{Competition results}
 The official results and winners were announced in mid-October 2021 and presented at a webinar in February 2022\footnote{L2RPN Webinar: \url{https://www.youtube.com/watch?v=WOt8xgpC370}}. The best performing agents are the ones that achieved a combination of high operational and attention scores. %Those who achieved high operational score but with lower consideration for attention score such as Maze-rl and IndigoSix did not rank to the top.
 The results confirm that there was a sufficient incentive to take into account the trust aspects (issuing the right alarm at the right time) besides the pure operational performance (running the power network) and this validates the framework used for modelling and evaluating trust in the competition.  % to eventually win the competition. 
 Given that operational performance have already been analysed in depth in previous competitions, in the below sections, we focus on the trust aspect and the related attention score of the competition winners.

\begin{figure}[htbp]
\centering
\includegraphics[width=\linewidth]{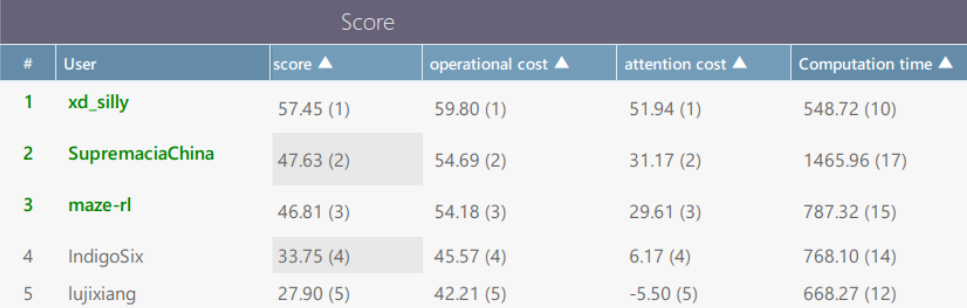}
\caption{Final ICAPS competition leaderboard }
\end{figure}

%The top 5 ranking agents all achieve pretty good operational performance. 

%Supremacia China is an advance expert system leveraging reduced action space of last year NeurIPS winner for best actions. For alarm, it is rule based above a threshold of max rho=1. It also make sure alarm don't overlap for at least 5 timestep difference
%xd_silly is a learning agent which simulate top sequences of up to 4 actions based on some predicted rho from a network. It uses thoses to predict also if he should send an alarm if no actions predict rhos below 1!

Analysing the results over the test scenarios, the two best agents Xd\_silly (\textcolor{blue}{Xd}) and SupremaciaChina (\textcolor{blue}{SC}) successfully operate the network (i.e. without the network blacking out) over 16 out of 24 scenarios. Overall, both agents have 7 failing scenarios in common. The best agent sends valid alarms in 5 out of 8 failing scenarios and the second best in 4 out of 8.i.e. predicted failure rates of 63\% and 50\% respectively.  
In these scenarios, the agent  sends alarms from 3 to 7 timesteps ahead of the failure (7 being the calibrated ideal time of alarm), which might be an indication of its planning time horizon. When sending alarms beyond 7 timesteps, it was usually the case that the operator's attention budget had diminished (so there was previous instance of many alarms) so it could be concluded that the agent would probably have resent alarms later on if the attention budget was sufficient. 
%for nov34_1, no more budget for xd_silly
%for dec12_2 no more budget either for both
%for jan_28_1, same for supChina
% same for jan_28_2
%it learnt to send alarms every 3 to 4 timesteps when in real difficulty

\begin{table}[H]
    \centering 
    \caption{Best alarm time and score comparison over failing scenarios for the 2 best agents}
\begin{tabular}{|>{\columncolor[gray]{0.9}}c||c|c|c||c|c|c|}

    \hline
    \multicolumn{1}{l}{\textbf{Scenario}} &
    \multicolumn{3}{c}{\textbf{Xd\_silly}} &
    \multicolumn{3}{r}{\textbf{SupremaciaChina}}
    \\
    \hline
    \rowcolor{lightgray}
      & $\bar{S}$ & $t_a - \bar{t}$ & $\bar{t}$ & $\bar{S}$ & $t_a - \bar{t}$ & $\bar{t}$ \\
    \hline
    $dec12_1$ & \cellcolor{red!25}-200 & -2  & 66 & \cellcolor{red!25}-200& -2 & 66\\
    \hline
    $dec12_2$ & 56 & -9 & 710 & 64 & -8  & 709\\
    \hline
    $feb40_1$ & \cellcolor{red!25}-200 & -2  & 22 & 24 & -3  & 23\\
    \hline
    $jan28_1$ & 42.7 & -4  & 1997 &  \cellcolor{red!25}-200 & -15  & 790\\
    \hline
    $jan28_2$ & 66.7 & -7  & 678  & 56  & -9  & 668\\
    \hline
    $jun01_1$ & 100 & -7 & 953 & - &  - & 2016\\
    \hline
    $mar07_1$ & - & - & 2016 &  \cellcolor{red!25}-200 & -2  & 1700\\
    \hline
    $nov34_1$ & 64 & -8 & 1282 &  \cellcolor{red!25}-200 & -2  & 1267\\
    \hline
    $nov34_2$ & \cellcolor{red!25}-200 & -2  & 163  & 42.7 & -4  & 1656 \\
    \hline
  \end{tabular}
  \label{tab:results}
\end{table}

%for dec12_2, they get the timing right but are not accurate enough on the location
%for jan28_2 rather right timing 45 minutes ahead while doing no actions. Rater know what it is doing. But gives 2 zones, while one is expected. Same for xd_silly after
%for jan28_2, supremaciaChina is sending an alarm much too early and then doing a lot of actions getting distant from the reference topology. It seems kind of lost. xd_silly survives and fail later after a strong attack. But send several areas, while one is expected.

Looking in more depth at some statistics from the competition results, shows that \textcolor{blue}{Xd} requires less attention from the operator than \textcolor{blue}{SC}, and is also more cautious with its attention budget. Indeed, it sends about 0.63 alarm per day on average (compared to SC: 0.78), keeps an average budget of 2.5 (compared to SC: 2.2) and only spends 1.5\% of the time with an attention budget below 1 (compared to SC: 10\% of the time). This highlights that Xd has more advance behaviour in regard of its ability to warn an operator when it is most needed, possibly suggesting a better assessment in the confidence of its actions and capabilities. In terms of actions, \textcolor{blue}{Xd} performs also less actions compared to SC, both on average per week (23.5 versus 26.5) and at maximum (38 versus 64). It shows that Xd is somehow more efficient in its decisions and actions. We will now give a short description of the nature of those agents that could explain those observations.

\subsection{Description of best competition agents}
Both agents leverage the actions that were learnt by the best winning solution of NeurIPS 2020 L2RPN competition \cite{zhou2021action}. \textcolor{blue}{Xd} is a hybrid agent that combines learnt modules and simulation. One learnt module based on a Deep Neural Network gives fast predictions of action impacts on line powerflow margins. They use this predictive model to explore the best possible combination of actions up to a depth of 4, defining a planning horizon over 4 timesteps, but without explicitly taking uncertainties into account over this horizon. They further simulate the top candidate sequences. Thanks to this feature, an alarm is not naively raised as soon as an overload appears, in the case when a sequence of actions is expected to relieve it. If none has been found to relieve existing overloads, only then an alarm would be raised. A rule eventually prevents re-sending alarm in less than 3 timesteps. 

The second best competitor, \textcolor{blue}{SC}, is an advanced expert agent, which makes proper use of rules and simulation to select the right actions in real-time over an initially curated database of effective actions. It however does not build a planning horizon and is closer to a greedy agent in that regard. Its alarm module is, in part, rule based, checking if overloads exist, if some lines are off and letting at least 5 timestep interval between alarms. It is nevertheless quite reactive for any overload showing up and could be quick at depleting its attention budget as we have noticed before. An additional alarm model is learnt to predict a percentage of how appropriate sending an alarm is at a given time. When above a threshold of 70\%, an alarm could be sent. These two strategies look complementary and it has been assessed on validation scenarios that this learnt model when combined with the rule-based one improves the attention scores by few points.

Given those characteristics, we will now make a more detailed behaviour analysis over some interesting scenarios.

\subsection{Behaviour analysis of competition agents}
%\antoine{make a connection with intimacy, reliability, explainability}
\begin{figure}[htbp]
\centering
\includegraphics[width=\linewidth]{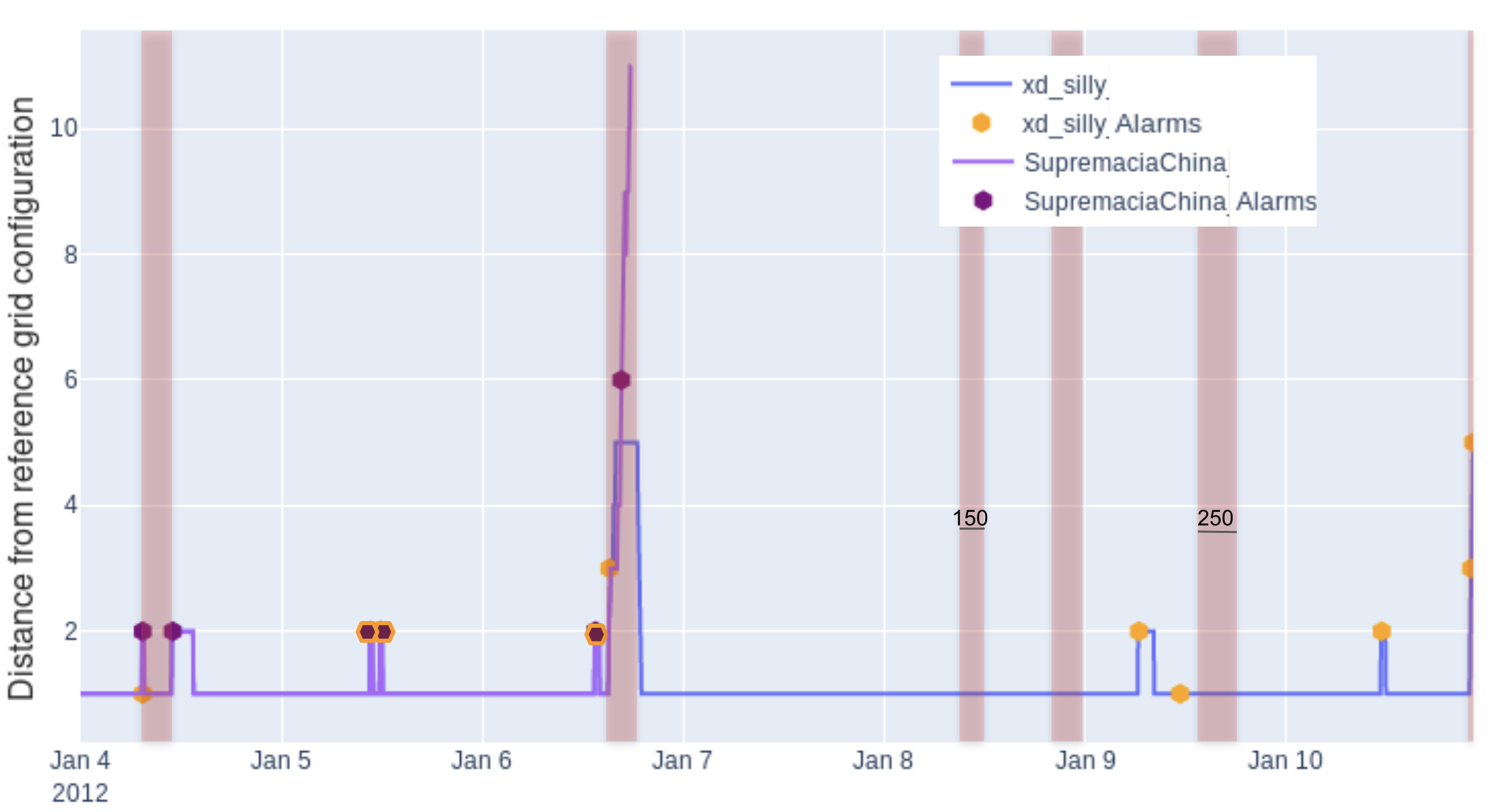}
\caption{Two best agents behaviour over time and before respective failures in scenario jan28\_1. It shows times of actions (as the topology distance varies) and alarms, and periods of attacks)}

\label{fig:fig_behaviour}
\end{figure}

From the list of scenarios that the winning agents failed to solve,
dec12\_2 and jan28\_2 are interesting for judging how well the two agents can anticipate its time of failure, instead of reacting and merely surviving attacks. Indeed, in theses cases, failure occurs in the last part of an attack period and not right at the beginning. The two agents have the last alarm timing right, about 7 timesteps before collapse, but are not accurate enough on the location (spatial area) of the failure. Based on this observation, we can assume that both agents have developed quite good prediction and planning skills over a fairly long time-horizon. But this hypothesis is nevertheless mitigated by the fact that they also run out of budget and would have probably sent one more alarms if they could have. They somehow "luckily" run out of budget at the right time. 

A similar situation for the \textcolor{blue}{SC} agent occurs early evening of January 6th in scenario jan28\_1 as shown in Figure \ref{fig:fig_behaviour}. But this time it sends its last alarm too early (more than one hour ahead) before running out of budget. It then runs into a long sequence of actions, deviating strongly from the reference topology. In this sequence, it seems that the agent does not know what it was actually doing and where it was going: a characteristic of a greedy behaviour as we have suspected before. At that point, \textcolor{blue}{Xd} survives with a (somehow smart) sequence of 4 actions. In this scenario, we also see that agents often send alarms during periods of attacks but not only during these periods.The \textcolor{blue}{Xd} agent almost entirely manages this scenario but eventually fails. It nonetheless managed to save enough attention budget to send a proper alarm before failing, thanks to not being too eager at sending alarm and spending its attention budget as seen in the statistics. When considering how an agent would interact with a human in reality, this is good behaviour. It alerts the operator in time that a major event will occur without action, prompting them to take over system operation. We also notice this at the beginning of the scenario: it only sends one alarm instead of the two alarms sent by \textcolor{blue}{SC}.Sending two alarms risks over exercising the operator's attention.  

In the other failing scenarios, the reason why the agents get a penalty score, is that they send an alarm too late and are not able to survive long enough. This often happens right after a strong attack on one of the high voltage lines.

%\feb_40_1, mar_07_1,nov_34_1
%dec_12_1: Supremacia China has spent too much alarm budget before, sends an alarm too early and then don't hhave budget anaymore. But this says it is not confident about its actions. They Globally send alarms at the same time

Finally, in none of these scenarios did we notice a willingness to deliberately fail at a given timestep to possibly maximise the attention score after sending an alarm earlier. This is reassuring for the framework and competition design. %as this was a probable issue that was considered when designing and calibrating the competition, not to give bad incentives for an agent to end learning an undesired behaviour. 

Our two agents showed good \textbf{reliability} due to its good operational performance and an ability to raise alarms before failure. They however mostly failed on the \textbf{credibility} side, not being selective enough on the time of alarm and the area of failure. In terms of \textbf{intimacy}, they also appeared limited, not taking sufficient account of its bounded attention budget when sending alarms. 
All this suggests that they cannot yet be considered trustworthy enough. For such complex acting agents. Is this a limitation of rule-based alarms? Would it be necessary to learn an alarm model instead? We now try to give some insights to those questions through dedicated experiments.
%Agents always tried to survive, or sometimes just stopped doing action when things get out of reach such as on scenario dec12\_1 for best agent. 

%\subsection{Sending Alerts: experiments and challenges}
%\subsection{{\color{blue}Sending Alerts: Insights from controlled experiments for future research}} 
\subsection{Investigation of trust concept with example agents}
%Insights from controlled experiments for future research on trust concept
%Investigation of trust concept with example agents 

With example agents, we aim to gain insights into some remaining challenges when developing an agent with a successful alarm feature. To this end, the uncertainties of the power system operation and the constraints of sending meaningful alarms result in several challenges: 

1) given the attention budget $\alpha_t$, the agent has to decide carefully when to send an alarm without wasting its budget, 

2) To make the alarm successful, the agent has to send it in a particular time window before the failure/collapse (defined as `game over'). As the underlying environment is stochastic (\textit{eg} random possibilities of lines being attacked) it is often too difficult to precisely predict the `game over'. 

3) On the other hand, an agent's successful alarm sending capability is directly linked to its current action. Hence, the challenges in designing the alarm feature increase with the increase in complexity of the agent's action selection criteria.

Next, we investigate in detail the design of agents with alarm features. To ease our understanding, without loss of generality, we can conceptually split the agent into two distinct parts, a) action-making, b) alarm-generating. 

First, we try to design a simple rule-based alarm agent. As mentioned earlier, a sound alarm agent can detect a possible danger in the running condition of the system. To this end, the most obvious choice is to monitor the capacity of each power line $\rho$, which is defined as the observed current flow divided by the thermal limit of each power line. Besides, there are possibilities that a power line can be attacked or can be disconnected due to maintenance, and any line disconnection obviously stresses the system operation. Hence, we extract the necessary information from the current observation and define a simple rule-based alarm feature agent (\textcolor{blue}{RbA-I}) as given in \textbf{Algorithm-1}. 
\begin{algorithm}[!htbp]
	\caption{Rule-based Alarm Agent-I} 
	\begin{algorithmic}[1]\label{algo1}
        \State Check whether any line is disconnected or attacked.
	    \If {disconnection or attack}
	    \State Check for any overload: 
	    %\vspace{-0.1in}
	    %\begin{align*}
	    %\underset{{l\in \textbf{all line}}}{\max}\;{\rho_l} > 1.00 
	    %\end{align*}
	    \If {Overload} 
	    \State Detect zones of overload and send an alarm.
	    \EndIf 
	    \Else 
	    \State Do not send any alarm.
        \EndIf 
	\end{algorithmic} 
\end{algorithm}
The design concept of this alarm feature is straightforward, and we tested this feature with two different action-making agents i) `Do-Nothing Action Agent' (\textcolor{blue}{DN}) and ii) `Simulation-intensive Expert Action Agent (\textcolor{blue}{SiE})'. In two different instances of testing, we observed that %in (a) Seed No-1: `
\textcolor{blue}{DN} + \textcolor{blue}{RbA-I} can send 14 successful alarm out of 24 different monthly scenarios. While \textcolor{blue}{SiE} + \textcolor{blue}{RbA-I} sends 10 successful alarms out of the same 24 different scenarios. In this testing phase, we %considered that each episode is of 4 weeks long, and it is
observed that no scenarios are completed till the end by any of the agent. %{\color{blue}, this is done with scoreicaps2021 function, and master seed from 'val'}. 
%(b) Seed No-2: In another phase of testing, we considered 1 week-long scenarios. Here, we observed that `Simulation-intensive Expert Action Agent + Rule-based Alarm Agent-I' successfully completes 8 scenarios and fails in 16 scenarios, while `Do-Nothing Action Agent + Rule-based Alarm Agent-I' fails in all 24 scenarios. Next, we study whether the agents can send successful alarms in their failures. It is found that the success rate of `Do-Nothing Action Agent + Rule-based Alarm Agent-I' in terms of sending alarm is $50\%$, while in the case of `Simulation-intensive Expert Action Agent + Rule-based Alarm Agent-I', the success rate is $37.5\%$. 
We can state that the simple rule-based alarm feature can be good for \textcolor{blue}{DN}, but the same is not as suitable for complex action agents. The reason is quite apparent; in \textcolor{blue}{DN}, the agent does not take any corrective action. Thus, it can be easily inferred that when the system is operating with one or more line outages and at the same time this power system is overloaded, failure is inevitable. In contrast, an \textcolor{blue}{SiE} can solve some difficulties after executing necessary corrective actions. The simple alarm agent fails to interpret the outcomes of the expert actions and sends unnecessary alarms thinking that there is an impending collapse. This ultimately reduces their attention budgets, makes them unable to send an alarm when the situation needs it. Plus, the operating conditions of failure for particular scenarios are not the same in the case of  \textcolor{blue}{DN} and \textcolor{blue}{SiE}. Hence, there may be the possibility that the \textcolor{blue}{DN} fails for simple reasons that are easily detectable. While the failure of \textcolor{blue}{SiE} is due to some complex reasons, the simple alarm agent fails to detect the same. This implies that the alarm feature of the agent needs some improvement, in order to perform well with a complex action agent. To improve the alarm feature, we studied some of the failures with unsuccessful alarms. It is found that attention budget and the timing of the alarm are playing key roles. Mostly, the alarms are sent but are not successful because either (a) the agent does not have the required amount of budget to send a successful alarm, or (b) the collapse occurred suddenly after a line attack, hence the alarm did not meet the desired time-window requirement. To tackle such situations, the agents need to predict the outcome of a line attack before the attack actually happened. Here, we modify the alarm features given in \textbf{Algorithm-1} and add some additional condition for sending alarm defining \textcolor{blue}{RbA-II} agent:  %if the predicted outcome of any attack breaches some predefined threshold $T_h$. The rules are as follows:
\begin{itemize}
    %\item[(i)] Check whether any there is any overflow, $\underset{{l\in \textbf{all line}}}{\max}\;{\rho_l} > 1.00$, if there is any overflow, generate alarms for the zones, where the overflow occurred.
    \item Simulate N-1 for the attacked lines list. If an overflow is predicted and $\underset{{l\in \textbf{all line}}}{\max}\;{\rho^{\text{pred}}_l} > T_h$, and there is no alarm in last $D$ time-steps, generate alarms for the zone where the predicted overflows exceed the defined threshold $T_h$.  
\end{itemize}
With this modification, the same set of scenarios 
%in (a) Seed No-1
: \textcolor{blue}{DN} + \textcolor{blue}{RbA-II} and \textcolor{blue}{SiE} + \textcolor{blue}{RbA-II} sends respectively, 21 (previously 14) and 13 (previously 10) successful alarms. 
This number increased from the one found with \textcolor{blue}{RbA-I}, especially for the \textcolor{blue}{DN} agent but there was a smaller increase for the  \textcolor{blue}{SiE} agent. We see that designing a complex rule-based alarm agent does not greatly improve this score on top of a complex acting agent. 
%But, in the case of the scenarios in (b) Seed No-2: No improvements are observed compared to the `Rule-based Alarm Agent-I'. It is clear that the performance of this alarm agent depends on the choice of predefined threshold $T_h$, and a predefined $T_h$ may not be suitable for different scenarios. Consequently, here comes the idea of introducing a learning-based alarm agent, which is our future research direction. Learning-based alarm agent can adaptively adjust $T_h$ extracting some underlying feature of interest. Plus, we noted that if we simulate the system considering the possibility of a line attack, the prediction is a possible overflow in most of the time steps. But the overflow only happens when there is an attack, the purpose of the threshold $t_h$ is to filter out some of those cases to save the attention budget. 
A rule-based alarm agent is not enough and we believe that this alarm prediction part can be improved with the help of a learning-based agent. This part should be further investigated in the future.
%Overall, the design of the best alarm agent is a complex optimisation problem, and hence, to improve the performance AI-based agent is indispensable. 

\section{Conclusion}\label{sec:conc}

On the journey towards creating trustworthy AI-based assistants for future network operators, we have proposed a trustworthy framework that builds on the reliability, credibility and intimacy model of trust, by explicitly considering the human operators' mental workload and capability of addressing issues when early warning and relevant network information are provided. Through the L2RPN with trust competition in 2021, we have  successfully designed a realistic active warning environment to experiment and evaluate trust between humans and agents. Winning teams have achieved the best alarm scores overall, in combination with the best operational performance, and demonstrated good reliability. By relying mostly on rule-based alarms, there however remains room for improvement on the credibility and intimacy aspects. 
Learning based alarm agents could help address in the future these open questions.
%These innovations will help build trust between the human and agent by 1) Attracting the operator's attention 2) Directing the operator's attention. 
\thanksto{\noindent Submitted to the 22nd Power Systems Computation Conference (PSCC 2022).}

\bibliographystyle{abbrv}
\bibliography{references.bib}

% that's all folks
\end{document}